\pdfoutput=1

\documentclass[letterpaper, 10pt, conference]{ieeeconf}  %

\IEEEoverridecommandlockouts                              %

\makeatletter
\let\NAT@parse\undefined
\makeatother

\usepackage[utf8]{inputenc}
\usepackage[comma,numbers, compress]{natbib}

\usepackage{url}
\usepackage{graphicx}
\usepackage[usenames,dvipsnames]{xcolor}
\usepackage[final]{fixme}
\fxsetup{theme=color}

\usepackage{amsmath}
\usepackage{amssymb}
\usepackage{textcomp}

\usepackage{booktabs}
\usepackage{threeparttable}
\usepackage{multirow}

\usepackage[capitalize]{cleveref}

\usepackage{tikz}
\usetikzlibrary{arrows}
\usetikzlibrary{positioning,calc}
\usetikzlibrary{decorations.pathreplacing}
\usetikzlibrary{decorations.markings}
\usetikzlibrary{fit}
\usetikzlibrary{shapes.callouts}
\usetikzlibrary{shapes.geometric}
\usetikzlibrary{matrix}

\usepackage{pgfplots}
\pgfplotsset{compat=1.9}
\usepgfplotslibrary{groupplots}
\usepgfplotslibrary{units}

\usepackage{blindtext}
\usepackage{placeins}

\usepackage{adjustbox}
\usepackage{float}

\title{\LARGE \bf
Robust 6D Object Pose Estimation in Cluttered Scenes \\ using Semantic Segmentation and Pose Regression Networks
}

\author{Arul Selvam Periyasamy, Max Schwarz, and Sven Behnke%
\thanks{All authors are with Autonomous Intelligent Systems, University of Bonn, Germany. {\tt\small periyasa@ais.uni-bonn.de}}%
}

\usepackage{eso-pic}

\AtBeginDocument{\AddToShipoutPictureFG*{\AtTextUpperLeft{\put(0,\LenToUnit{9pt}){\parbox{\textwidth}{\centering\bfseries
International Conference on Intelligent Robots and Systems (IROS), Madrid, Spain, 2018
}}}}}
\begin{document}

\maketitle
\thispagestyle{empty}
\pagestyle{empty}

\begin{abstract}

Object pose estimation is a crucial prerequisite for robots to perform autonomous manipulation in clutter.
Real-world bin-picking settings such as warehouses present additional challenges, e.g., new objects are added constantly.
Most of the existing object pose estimation methods assume that 3D models of the objects is available beforehand.
We present a pipeline that requires minimal human intervention and circumvents the reliance on the availability of 3D models by a fast data acquisition method and a synthetic data generation procedure.
This work builds on previous work on semantic segmentation of cluttered bin-picking scenes to isolate individual objects in clutter.
An additional network is trained on synthetic scenes to estimate object poses from a cropped object-centered encoding extracted from the segmentation results.
The proposed method is evaluated on a synthetic validation dataset and cluttered real-world scenes.

\end{abstract}

\section{Introduction}

Many robotic applications depend on the robust estimation of the object poses.
Different tasks may, however, place varying emphasis on certain aspects, such
as preciseness, robustness, speed, or fast adaption to new objects.
Robustness, in the sense of general applicability across different objects,
is difficult to obtain: Transparent or featureless objects pose challenges for many pose estimation methods that require valid depth measurements
and/or rely on a fused 3D model.

Our work is motivated by the Amazon Robotics Challenge 2017, which required
participants to manipulate objects in cluttered bin picking scenes.
In particular, fast and robust adaption to novel items was required, with teams
only having 30\,minutes to enroll 16 new objects to their systems.
In this context, leveraging high-quality object models is difficult.

\begin{figure}
  \centering
  \input{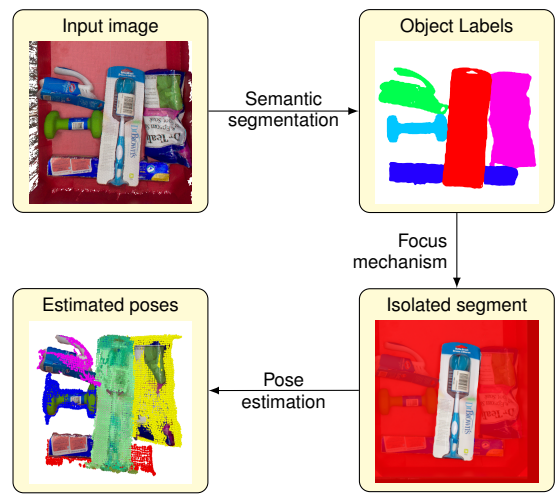}
  \caption{The proposed 6D pose estimation pipeline for cluttered scenes.}
  \label{fig:scene}
\end{figure}

We present a Convolutional Neural Network (CNN)-based regression pipeline for 6D pose estimation from RGB-D image segments,
which can be used following a semantic segmentation of the scene (see \cref{fig:scene}).
It first estimates a 5D pose from the RGB image, whose translation is then
projected into 3D using the depth information.

Our contributions include

\begin{enumerate}
 \item a fast object enrollment scheme for quickly learning new items from automated turntable captures,
 \item an encoding mechanism for focusing the network on a particular object
    in a cluttered scene,
 \item a CNN pipeline for direct pose regression, and
 \item explicit handling of object symmetries.
\end{enumerate}

\section{Related Work}

Traditionally, pose estimation is done using registration-based methods.
Given a good 3D model of the object and a clear separation from the background, Iterative Closest Point algorithms and
their extensions perform extremely well \citep{holz2015registration}.
In cases where depth measurements are unreliable or a 3D object model is hard to
obtain, more complex methods are required.

RGB-only methods perform more robustly in these settings.
For example, \citet{zhu2014single} extracted object silhouettes and matched these
against 3D models.

Hybrid learning-based methods used both RGB and depth information, adapting
to impreciseness or missing depth information. \Citet{schwarz2015rgb}
demonstrated object detection and pose estimation using pretrained CNN features.
However, only 1D pose (yaw angle) was estimated.

\Citet{krull2015learning} used random forests to generate an object probability
distribution as in semantic segmentation and 3D object coordinates corresponding
to each object for each pixel in the image.
Then for each detected object, in a Random Sample Consensus (RANSAC)-like approach, 3-tuples of pixels were randomly
selected and the affine transformation to their predicted 3D object coordinates was estimated.
This produced a set of hypotheses $H$
using the Metropolis algorithm. RGB and depth images were rendered from the 3D model
for each of the hypotheses.
The authors then trained a CNN to take the rendered and observed images as input
and output an energy value that is high if inputs are
similar and low when dissimilar.
Finally, the hypothesis with maximum energy was selected.
In contrast, our work needs one forward pass per segmented object, but
cannot take advantage of iterative refinement.

\Citet{koorobolink} trained a CNN to regress the pose of parts using multiple rendered views of CAD models.
The method needed depth information to pick a single part from a pile. We alleviate the need for depth information to isolate an object in the clutter by computing a semantic segmentation of the scene and performing the pose estimation only on a crop of the scene.

\Citet{wohlhart2015learning} trained a CNN not to estimate the pose of the objects directly but to compute a feature descriptor such that the Euclidean distance between the descriptors corresponding to the two poses are smaller if the poses are similar and larger when the poses are dissimilar.

\Citet{kendall2015posenet,kendall2017geometric} used CNNs to regress the 6D camera pose from a single RGB image in a large-scale outdoor setting.
Note that this inverse problem does not require an attention mechanism to
focus the estimation on a particular object.
Both \citet{koorobolink} and \citet{kendall2015posenet} encoded orientation as a quaternion as we do in our approach.

In contrast to other RGB-D methods, our method uses depth only during the
capturing process before training the model,
and for projecting the inferred 5D pose estimate to 6D.
This increases the robustness against missing depth and allows 5D pose estimation
on RGB-only images.

There are multiple publicly available datasets for benchmarking object pose estimation methods, Some include Washington RGB-D \citep{lai2011large}, SUN RGB-D  \citep{song2015sun}, and OccludedLinemod \citep{brachmann2014learning}. However,
these are often focused on a small set of objects, limiting generalizability.
In recent works, such as SceneNet RGB-D~\citep{mccormac2017scenenet}, we can see
a trend towards large-scale datasets covering all kinds of objects, similar to
ImageNet~\citep{deng2009imagenet} in the image classification domain.
In contrast, our work focuses on a particular robotic application with fast adaption to
a novel set of objects.

\section{6D Pose Estimation}

We propose a pose estimation network based on direct regression of a 5D pose, i.e.
2D translation in the image plane and 3D orientation. The 2D translation is later projected into 3D using depth information.

\subsection{Data Acquisition \& Training Scene Synthesis}
\label{sec:data_acq}

For many applications, it is important to quickly adapt to novel object classes.
This adaption does not only encompass training the pose regressor, but also capturing
the necessary training data. Many works ignore this step and instead require
high-fidelity 3D models as training input, which can be hard to acquire.

\begin{figure}
 \centering
 \input{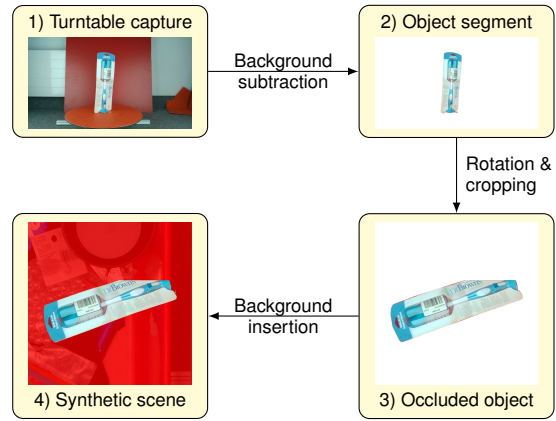}
 \caption{Data acquisition \& scene synthesis pipeline.}
 \label{fig:acquisition}
\end{figure}

In order to keep necessary human intervention minimal, we use an automated turntable
setup, which we designed for the Amazon Robotics Challenge
2017~(see \cite{schwarz2018fast} for details).
The turntable shown in \cref{fig:acquisition} captures 20 views from all directions in 10\,s with an
Intel RealSense SR-300 RGB-D sensor. We call this set of views \textit{sequence}.
For most of the objects three sequences are captured in different resting orientations on the turntable,
resulting in 60 views per object.
The object is segmented in each frame using background subtraction
(i.e. comparison with an object-less frame).

\begin{figure}
 \centering
 \includegraphics[height=0.3\linewidth,clip,trim=400 200 320 100]{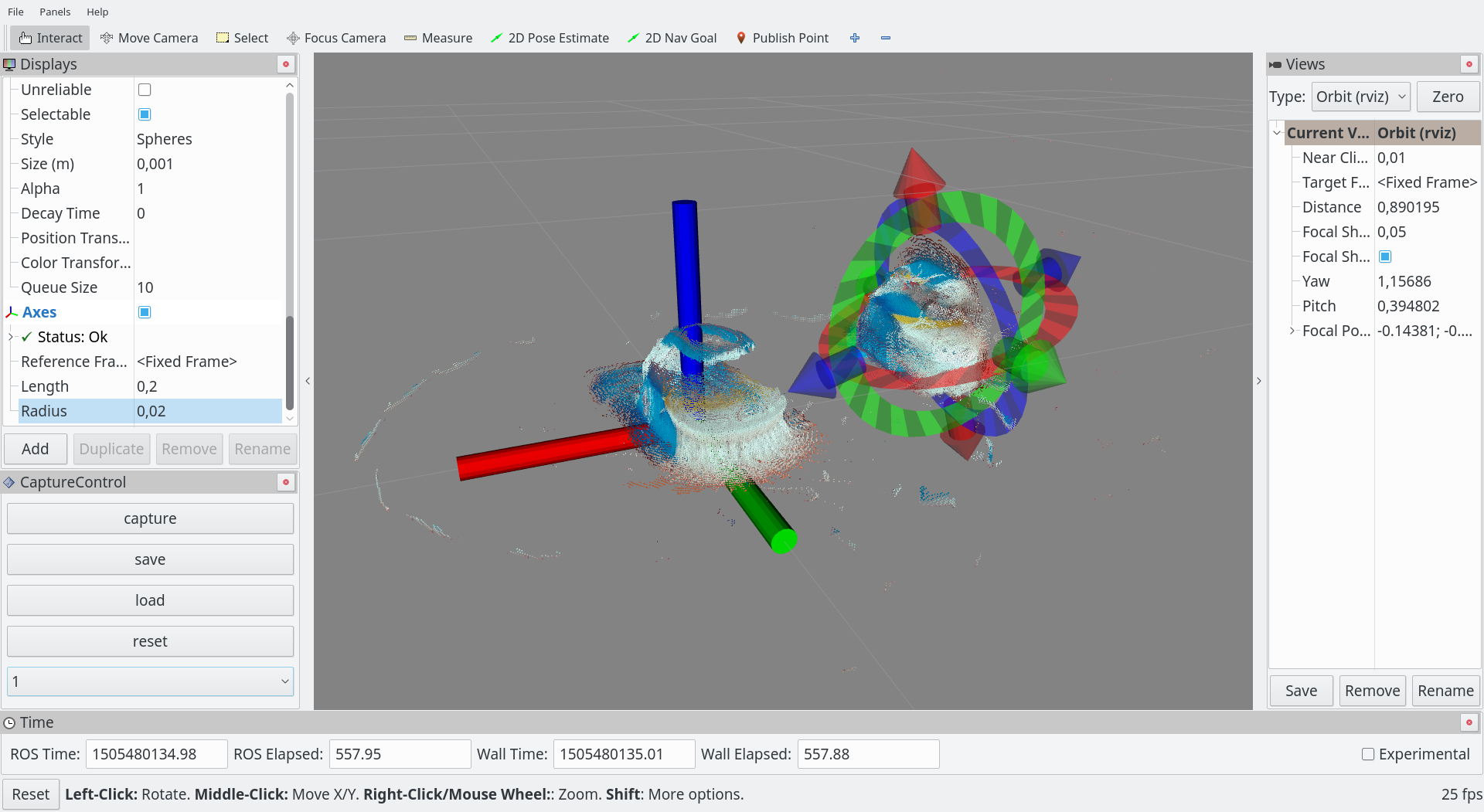}\hfill
 \includegraphics[height=0.3\linewidth,clip,trim=450 200 320 200]{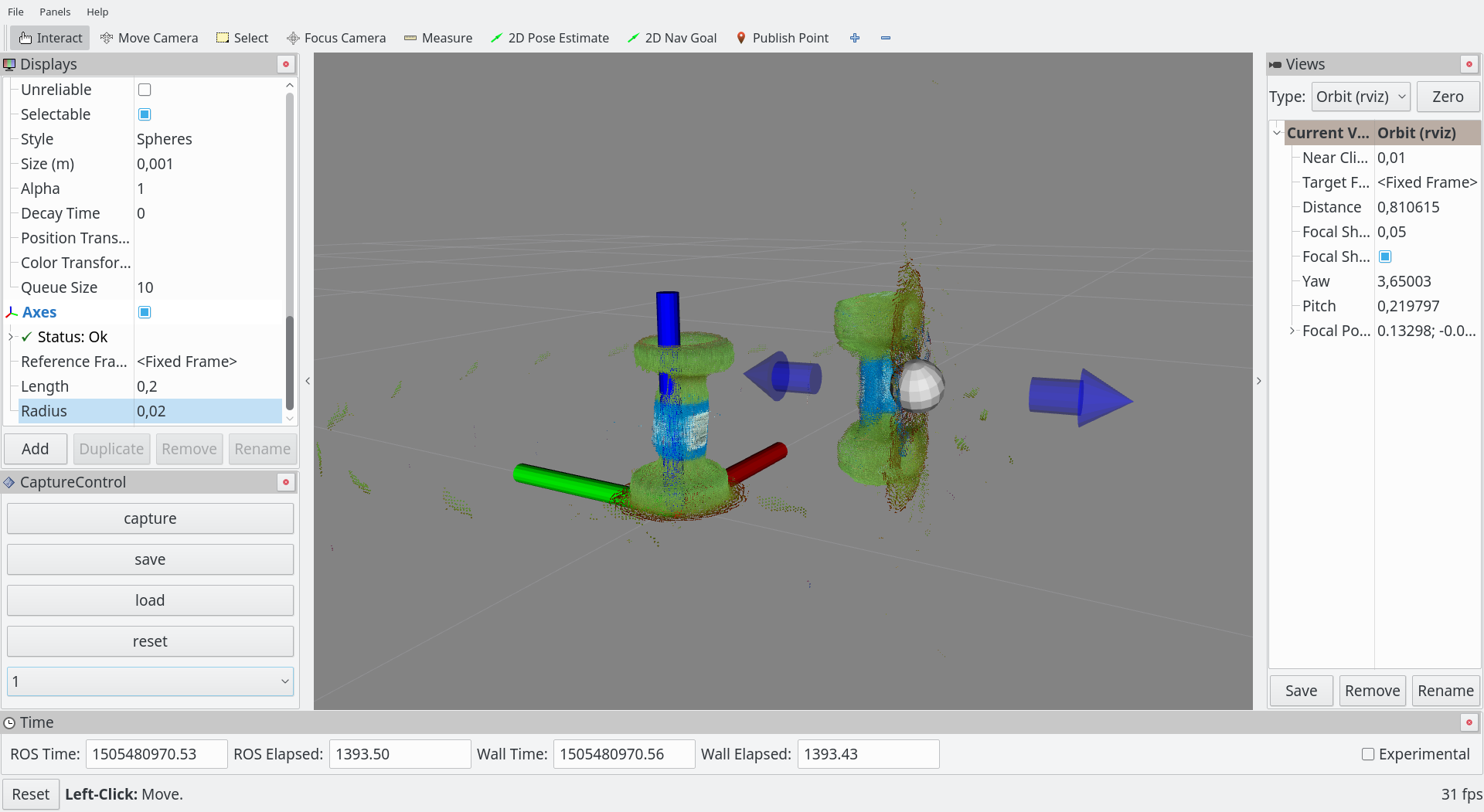}
 \caption[Alignment tool usage.]{Alignment tool usage. The coordinate system shows the frame of reference for the particular object.
 Already aligned sequence point clouds are shown around this origin.
 A new cloud is integrated by transforming it interactively using the 6D marker.}
 \label{fig:alignment}
\end{figure}

\begin{figure*}
 \centering
 \adjustbox{valign=m,raise=-2.5mm}{\includegraphics[width=.4\linewidth,clip,trim=0 0 0 0]{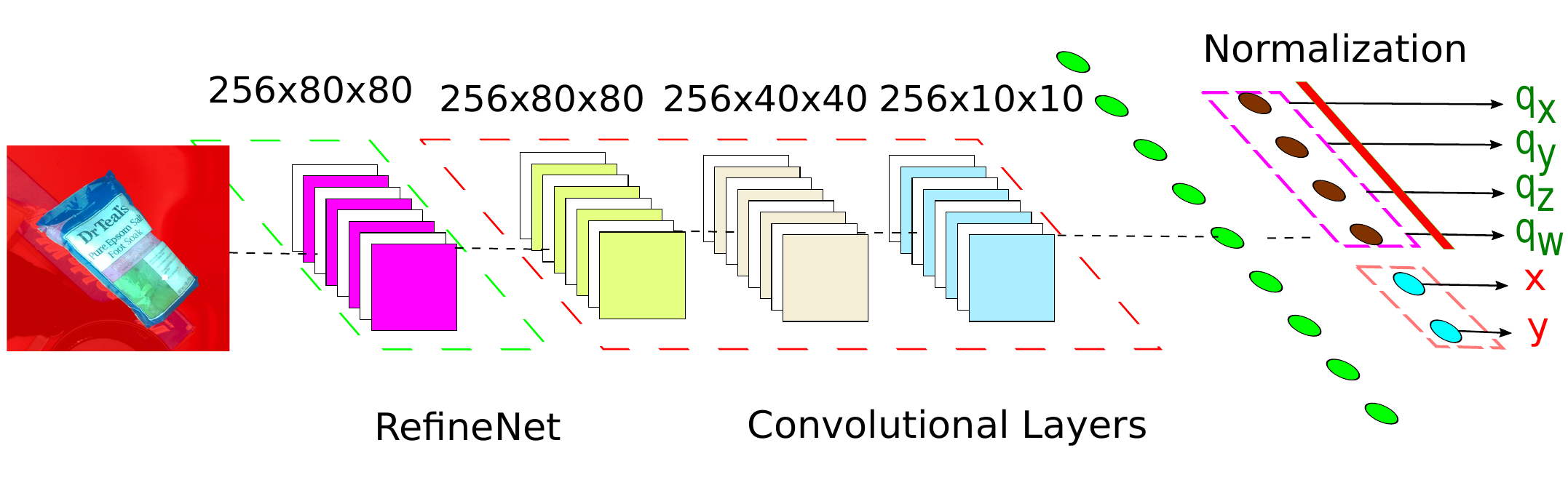}}\hfill
 \adjustbox{valign=m}{\includegraphics[width=.52\linewidth,clip,trim=0 0 0 0]{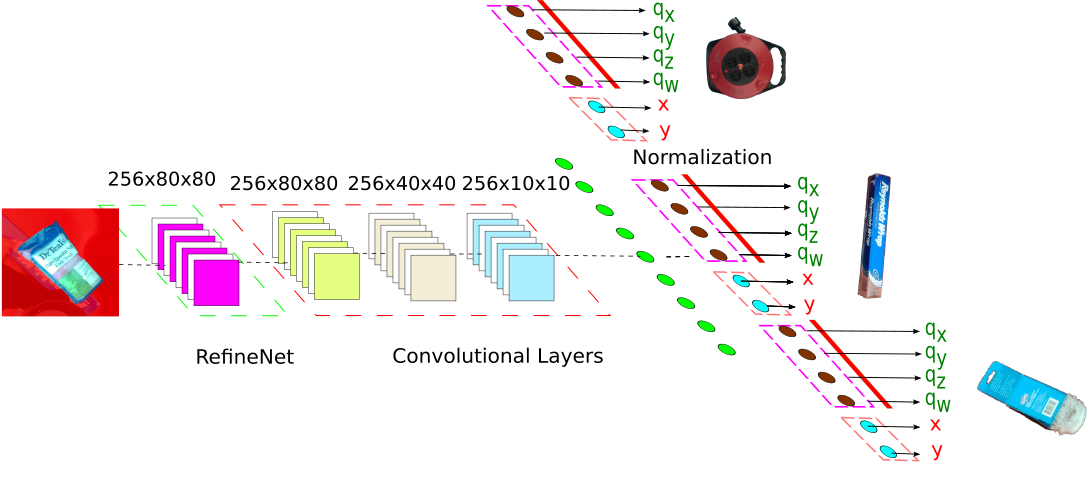}}
 \caption{Pose estimation network architectures. Left: Single-block output variant. Right: Multi-block output variant.}
 \label{fig:pose_architecture}
\end{figure*}

The turntable images are not immediately suitable for training, since they
show the object in an isolated setting without occlusions.
To address this issue, we introduce a scene synthesis step, which overlays the turntable images of objects
onto complex scenes. For generating training data for semantic segmentation, we start with images of complex scenes that were 
manually annotated beforehand. We overlay the turntable images onto it---occluding the already existing objects---and concurrently generate pixel-wise ground truth annotation.
For training the pose estimation, we just use the same background images (without annotation).
A special encoding scheme, discussed in the next subsection, is used to focus the network on a single object in a complex scene, eliminating
the need for background annotation.

The ground truth pose of the objects is generated using forward kinematics
from the measured turntable angle.
The alignment of poses between different sequences is done manually using an RViz-based\footnote{\url{http://wiki.ros.org/rviz}} GUI (see \cref{fig:alignment}).
Also, the available set of poses is augmented by sampling
random rotations around the camera axis.

\subsection{Encoding Of Object In Focus}
\label{encoding}
Since scenes may have multiple objects, we need a mechanism to
make the network focus on the object of interest. We encode the object of 
interest by pushing other pixels towards red (see \cref{fig:acquisition} Step: 4)).
This encoding is natural in the setting of our test dataset (see \cref{sec:dataset}),
which has red background behind the objects.
During training, this encoding is based on the background subtraction mask; during inference,
the semantic segmentation prediction is used.

\begin{figure}
 \centering
 \includegraphics[height=3.2cm,clip,trim={700 90 850 400}]{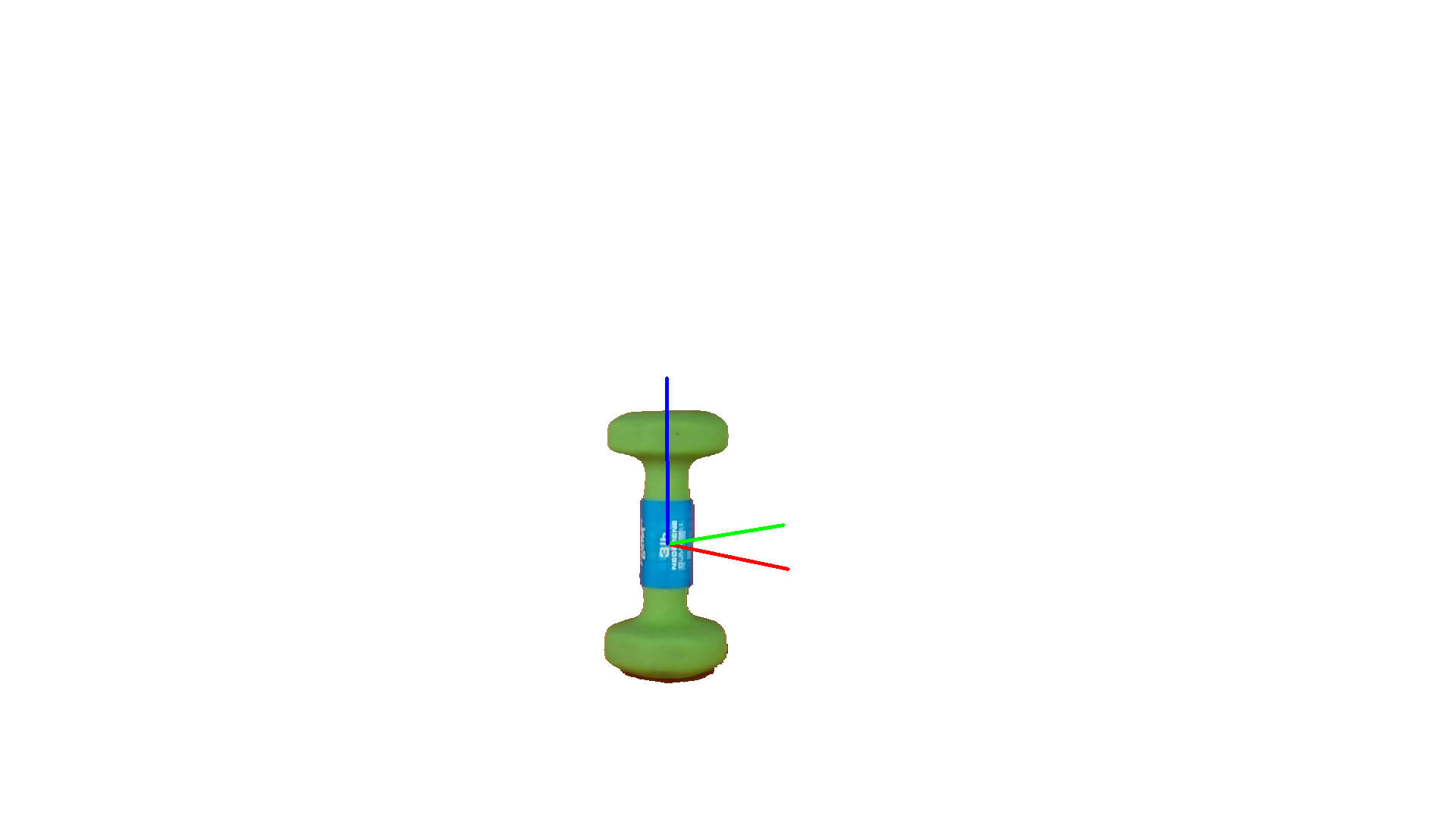}\hfill
 \includegraphics[height=3.2cm,clip,trim={700 90 850 400}]{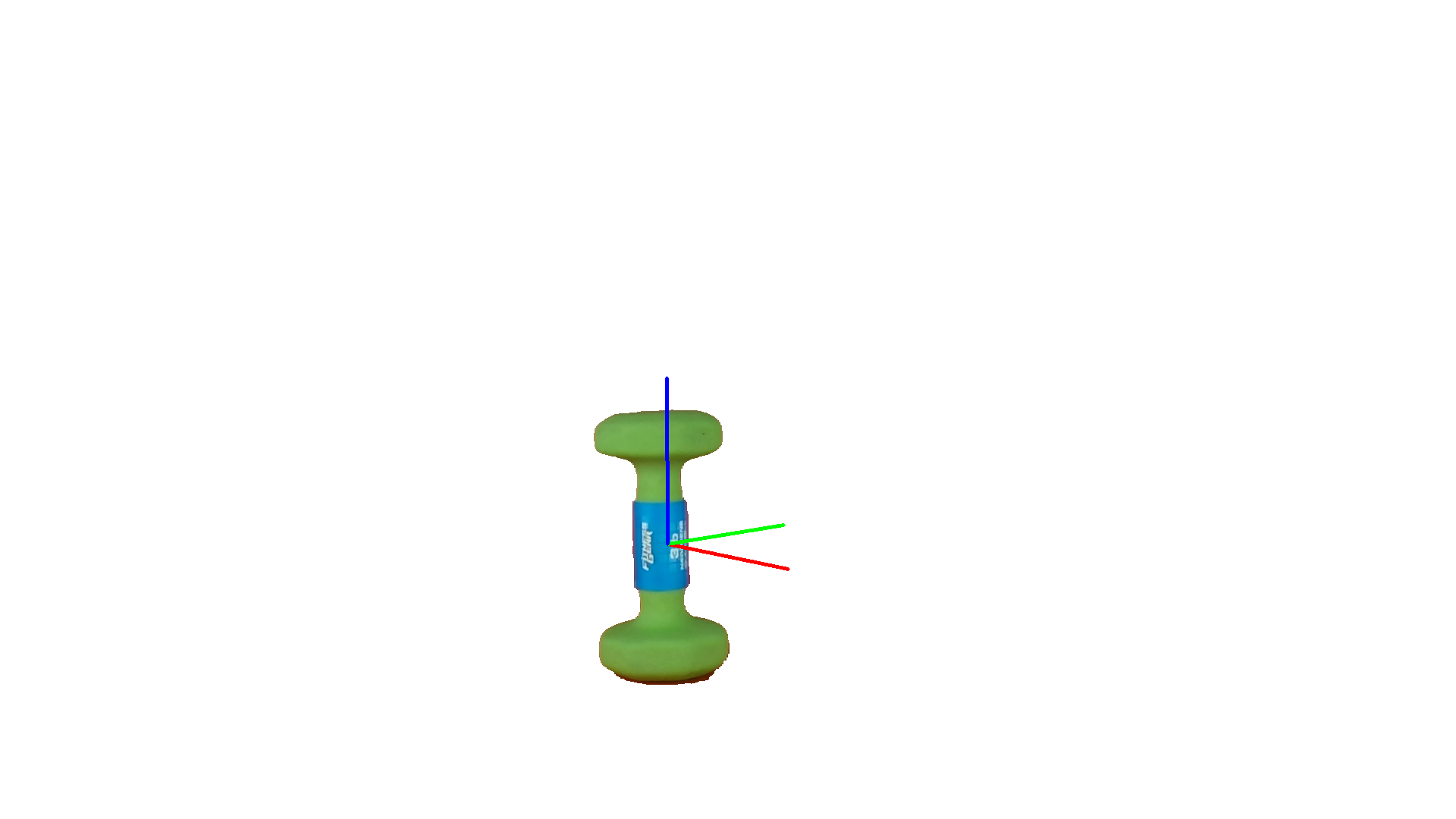}\hfill
 \includegraphics[height=3.2cm,clip,trim={700 90 850 400}]{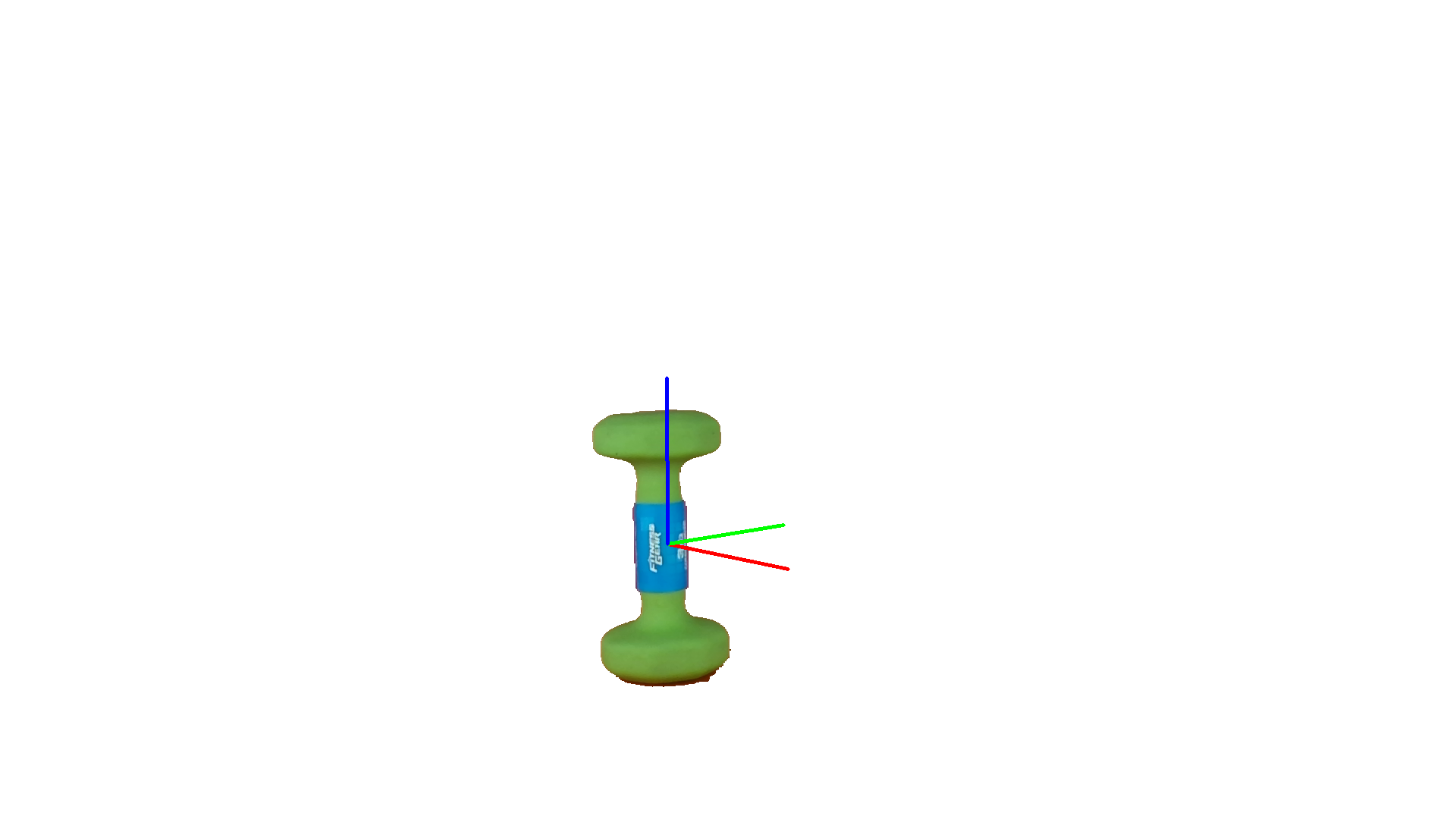}\hfill
 \includegraphics[height=3.2cm,clip,trim={700 90 850 400}]{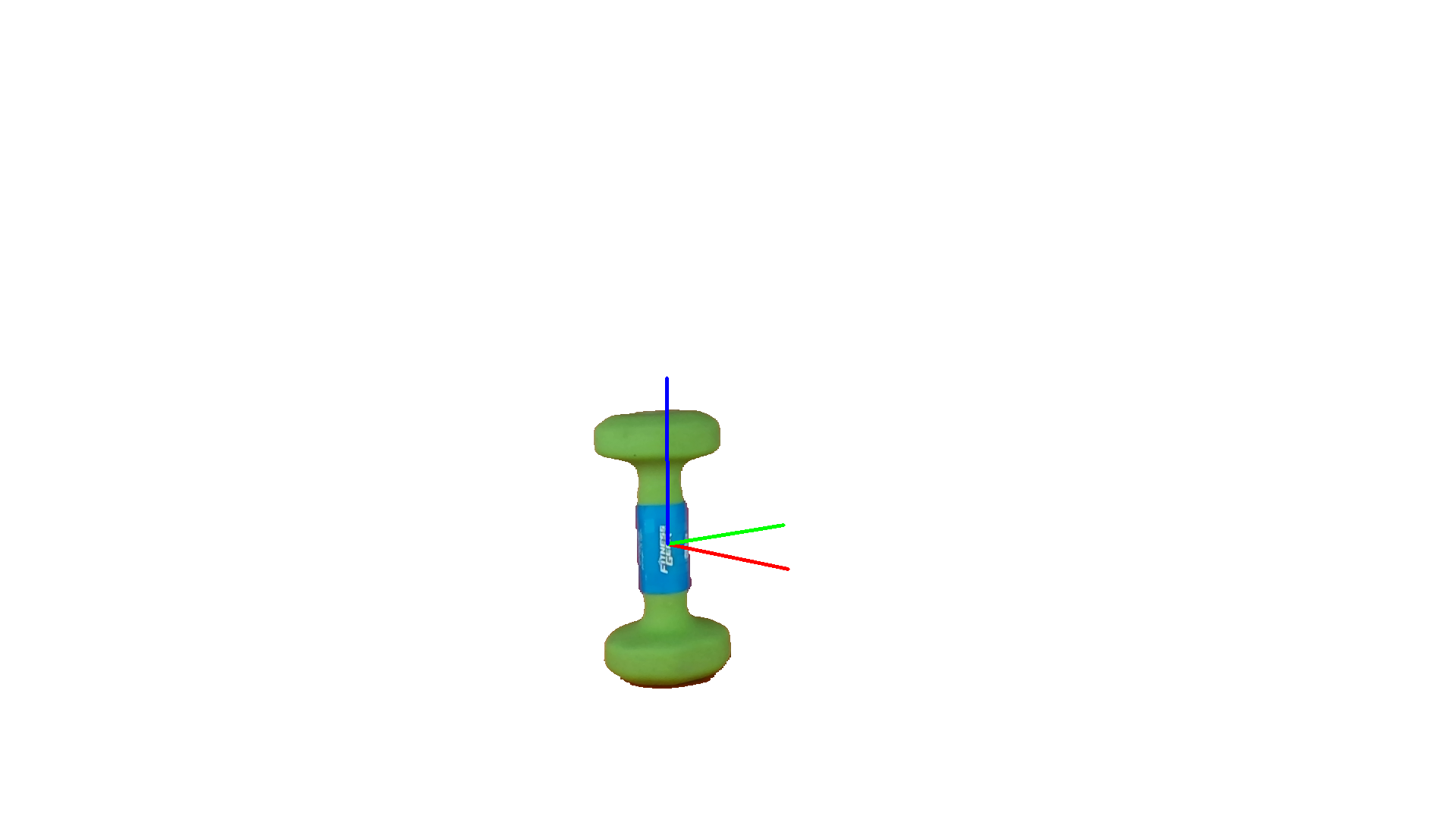}
 \caption{Resolution of symmetries. Invariant poses are assigned the same ground truth pose.}
 \label{fig:inv_ex}
\end{figure}

\begin{figure}
 \centering
 \includegraphics[height=4.0cm,clip,trim={700 150 850 170}]{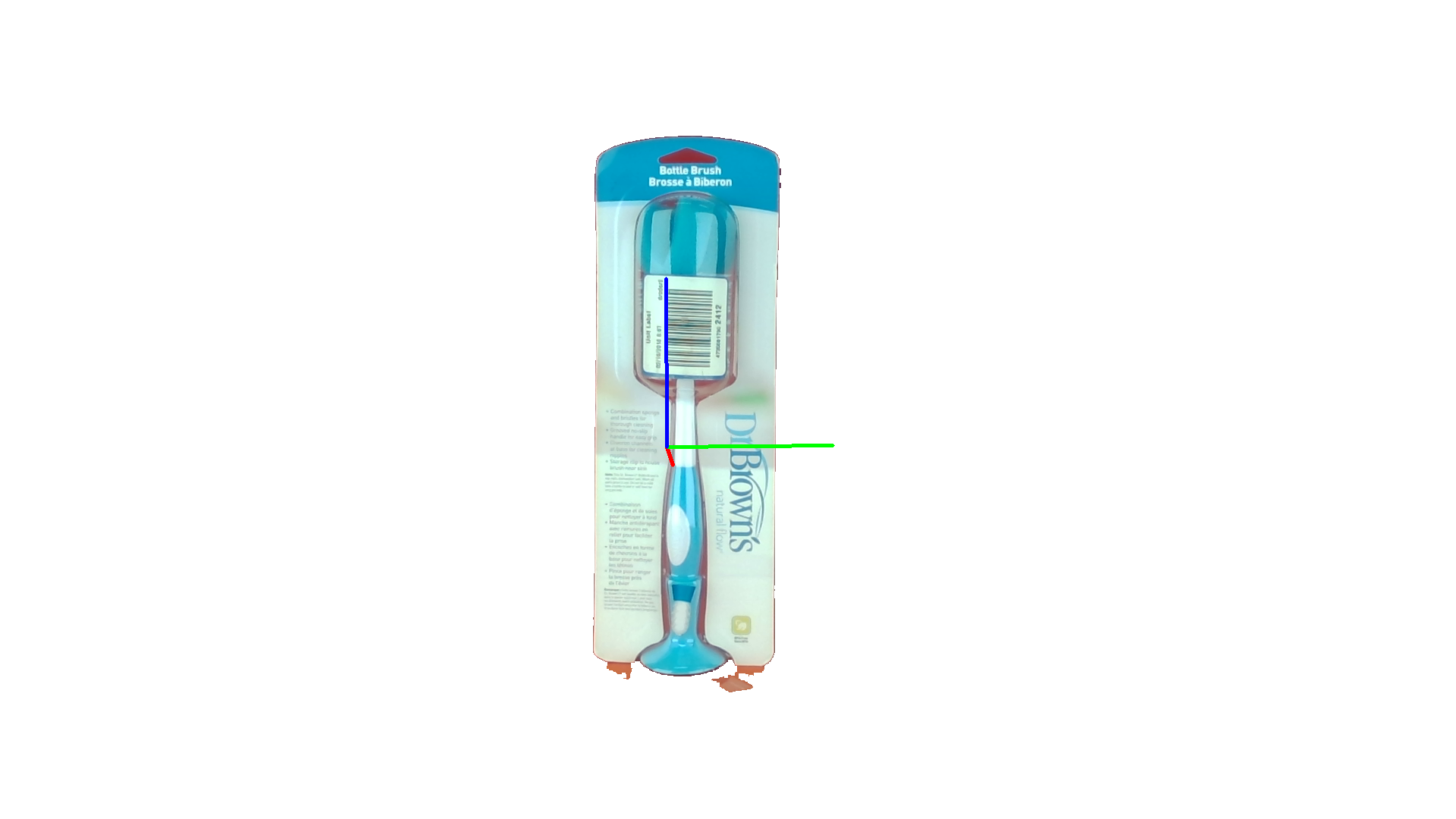}\hfill
 \includegraphics[height=4.0cm,clip,trim={700 150 850 300}]{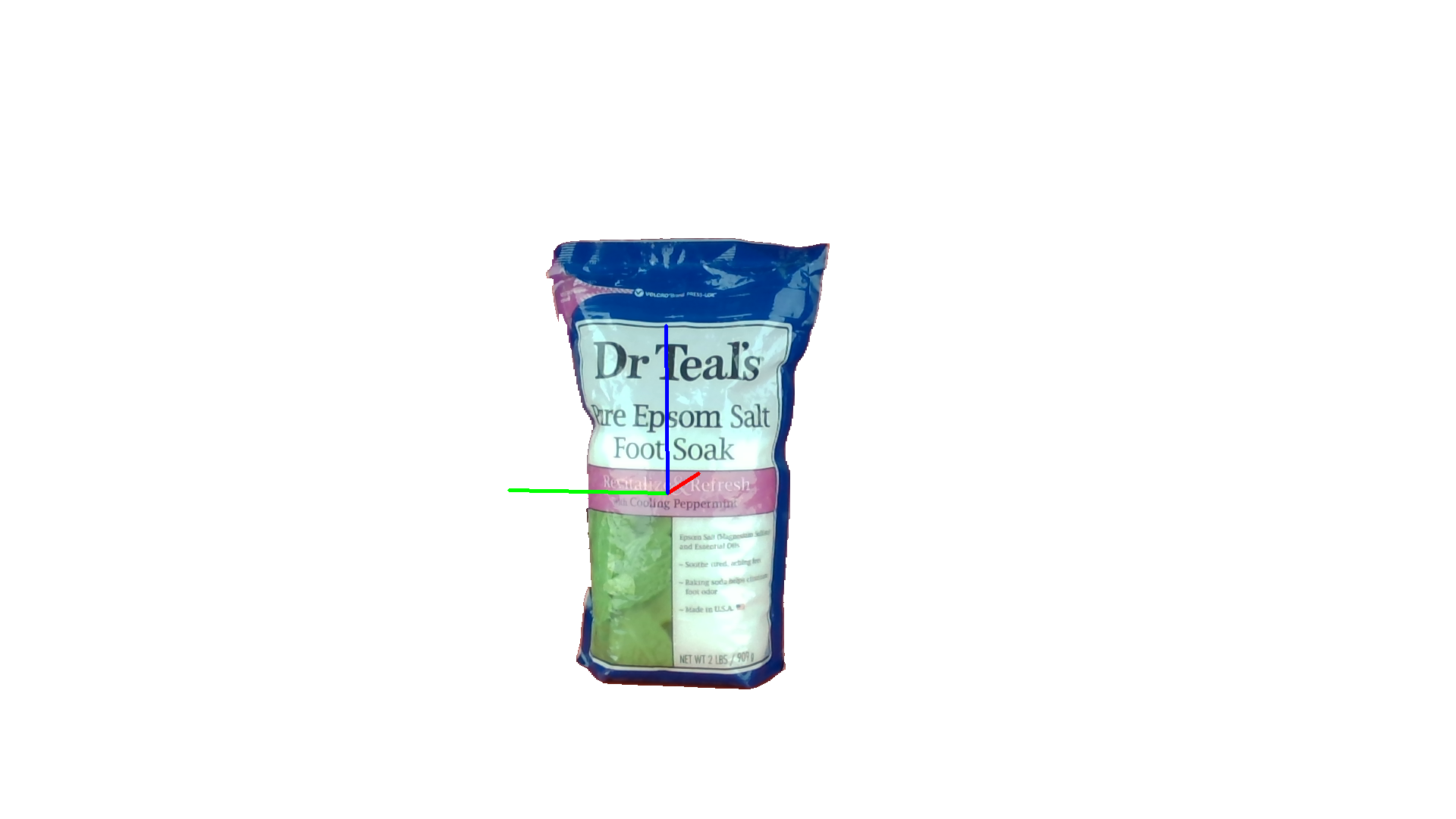}\hfill
 \includegraphics[height=4.0cm,clip,trim={700 90 850 100}] {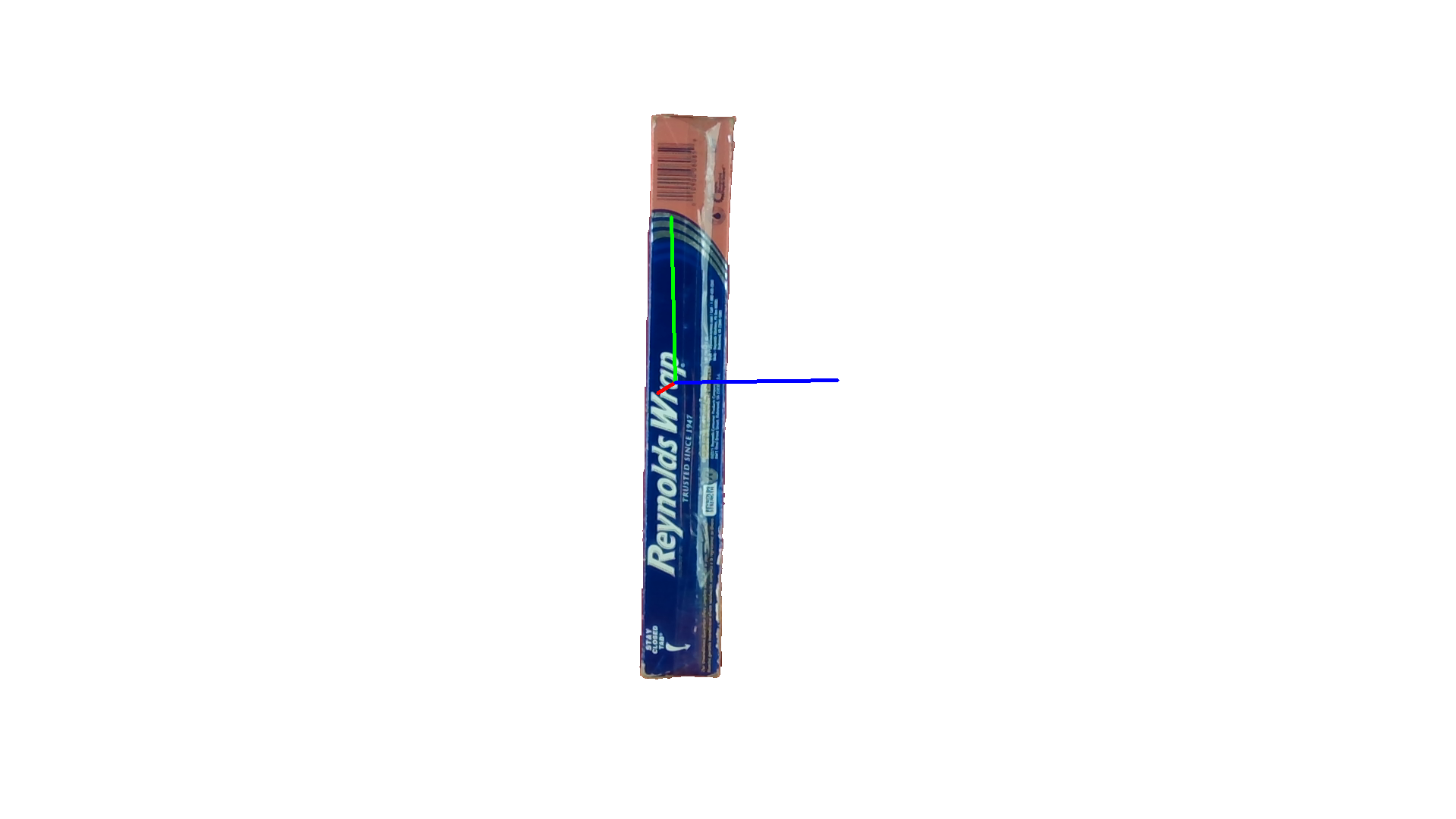}

 \includegraphics[angle=90,height=3cm,clip,trim={700 130 850 500}]{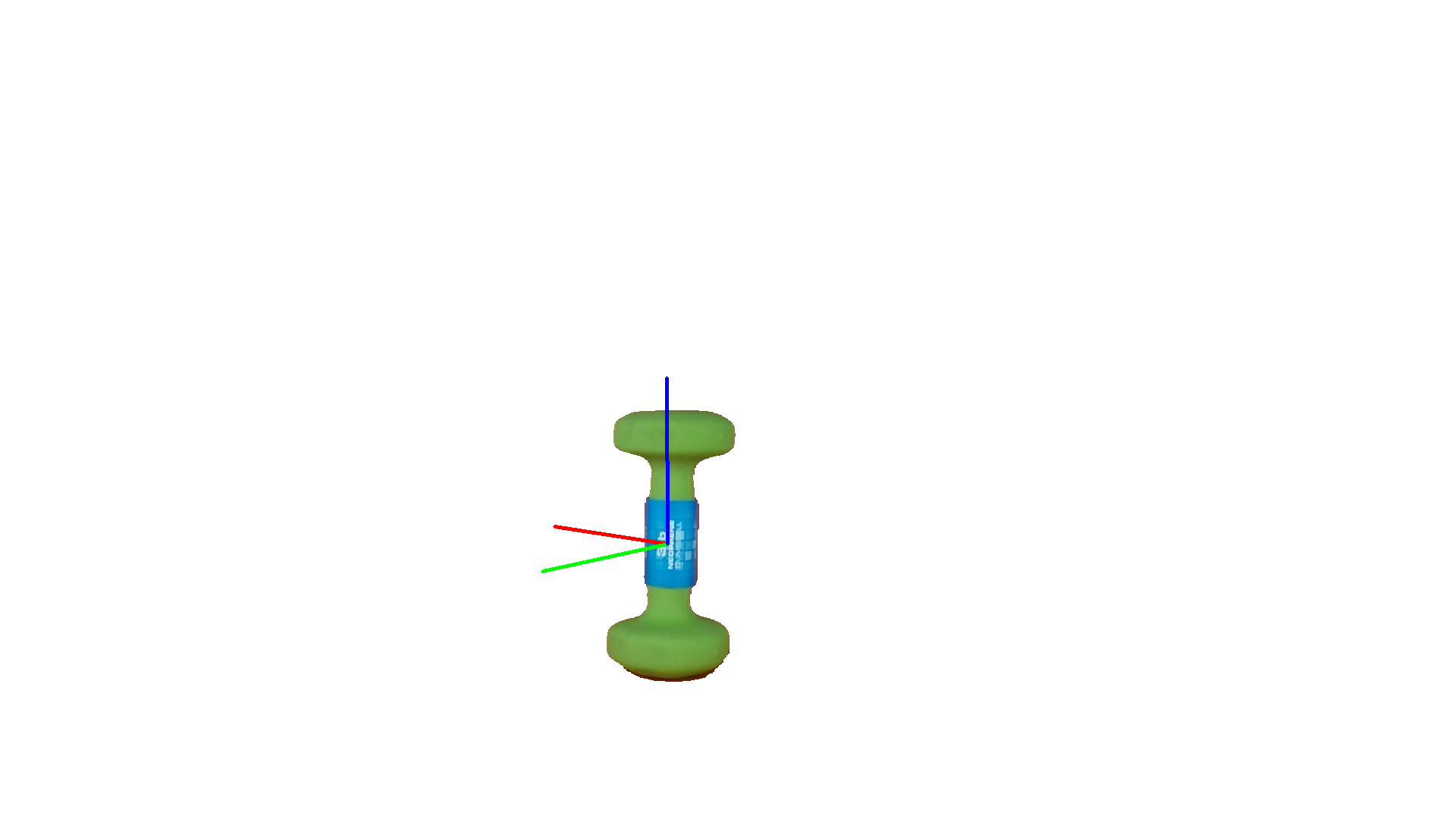}
 \includegraphics[height=3cm,clip,trim={600 90 700 600}] {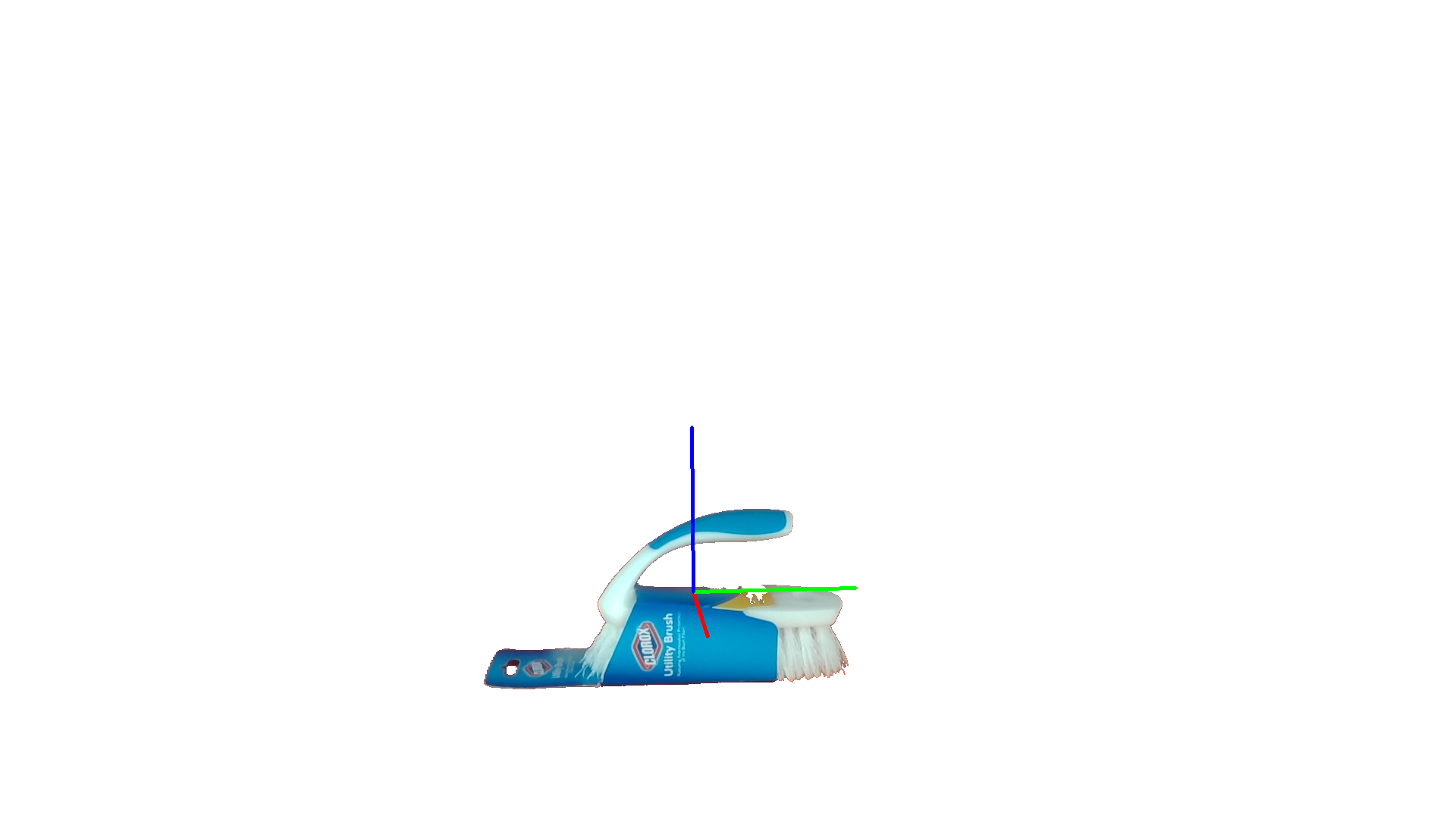}
  \caption[ARC objects used in pose estimation evaluation.]{
   ARC objects used in pose estimation evaluation.
   From left to right, top to bottom: Browns brush, epsom salts, reynolds wrap, hand weight, and utility brush.
   }
\label{fig:arc_objects}
\end{figure}

\subsection{CNN Backbone}

CNN-based methods can leverage pretraining from large-scale datasets.
Since these datasets often aim at the image categorization task (e.g. ImageNet~\citep{deng2009imagenet}),
the spatial resolution of the high-level features extracted by higher levels
of these CNNs is quite low.
In tasks such as semantic segmentation, this limits performance.

The RefineNet architecture~\citep{lin2017refinenet} mitigates this problem
by successively merging upsampled high-level feature maps with lower-level
representations, thus yielding a both highly semantic and spatially precise
result. In our work, we use the RefineNet architecture with
ResNeXt~\citep{xie2017aggregated} as the pretrained backbone network.
The RefineNet architecture is pretrained on the semantic segmentation task.

\subsection{Semantic Segmentation Network}
The semantic segmentation network~(see \cite{schwarz2018fast} for details) consists of RefineNet backbone followed by
one convolutional layer that reduces the number of feature maps to the number of object classes and pixel-wise SoftMax.
It is trained to minimize the pixel-wise cross-entropy segmentation loss on the synthetic dataset described in \cref{sec:data_acq}.

\subsection{Pose Estimation Network}

Our pose estimation network illustrated in  \cref{fig:pose_architecture} consists of three dimension-reducing
convolutional layers of kernel size 3 and stride 2 followed by two
fully-connected layers and six output neurons corresponding to the 2D 
translation in the image plane and the orientation represented as a
unit quaternion. All layers use ReLU activations.
Finally, an $L2$ normalization layer is added in the orientation part,
which guarantees a unit quaternion output,
alleviating the need for the network to learn the normalization.

Quaternions are not unique representations of rotation (i.e. $q$ is an equivalent rotation to $-q$).
To deal with this non-uniqueness in the representation, we require $q_w \geq 0$ for the ground-truth quaternion during the training phase.

For pose estimation, the scene is cropped to a size of 320$\times$320, centered at the origin of the object in the image plane.
Additionally, we randomly move the center of the crop to be a few pixels away from the origin of the object in the image plane.
Without this data augmentation, the objects in the training images always appear to be centered in the crop, and the model might learn 
to overfit this artificial condition.
During the inference step, where the ground truth origin of the object is not known, the center of the contour extracted from semantic segmentation is used.

Spatially, the backbone network reduces the 320$\times$320 input scene to
256 feature maps of size 80$\times$80 which are provided as input to the pose estimation network.

The network is trained to minimize the weighted sum of mean-squared-error (MSE)
of the translation and the orientation components.
The translation error $||\hat{Y}^{xy} - Y^{xy}||_2$ (pixel distance in a
320$\times$320 image) is of a different scale compared to the quaternion distance $||\hat{Y}^q - Y^{q}||_2$.
We scale the x and y translation to lie in $[-1,1]$,
which brings the translation and orientation errors into the same scale.
This allows us to use a simple weighting scheme for the loss components:
$$ \textrm{Loss}(\hat{Y}, Y) = \alpha||\hat{Y}^{xy} - Y^{xy}||_2^2
 + (1-\alpha)||\hat{Y}^q - Y^{q}||_2^2, $$
 where $\hat{Y_i}$ is the ground truth pose, $Y_i$ is the predicted pose for an image $i$, and $Y_i^{xy}$ and $Y_i^{q}$ are translation and quaternion component, respectively.

We empirically determined a value of 0.7 for $\alpha$. The network is
trained using the Adam optimizer \citep{kingma2014adam}.

\subsection{Multi-class Regression}

The pose network is trained to estimate the pose of objects belonging to different classes.
It is not immediately obvious how to handle this properly.
On the one hand, one can require the network to recognize the object class and output the
proper pose.
A second possibility is to predict \textit{conditional} poses, one for each
object class. An external module (for example semantic segmentation) then
picks the correct output for the detected class.

For the second case, we implemented a multi-block output variant of the pose estimation network, also shown in \cref{fig:pose_architecture}.
The network has $6N$ outputs for $N$ objects classes.
During training, the loss function is only applied to the outputs of the correct object class.
Both variants are evaluated in \cref{sec:eval:multiblock}.

\subsection{Symmetries}

Some objects may exhibit symmetrical appearance when rotated along a 
particular axis. For example, the dumbbell object shown in
\cref{fig:inv_ex} is symmetrical---with minor differences in the
text on the label---to rotation in yaw axis (in blue) and to a rotation of $180^\circ$ in roll or pitch (red and green).
Forcing the network to learn the exact pose from the images that exhibit very little
variance may hinder the learning process.
Also, learning the pose component that vanishes under (perceived) symmetry is not helpful for
robotic manipulation.
To deal with the symmetries in the object pose, we assign the same ground truth pose to all
poses that are symmetrical. \Cref{fig:inv_ex} shows an example.
The symmetry axes need to be specified manually during alignment.

\section{Evaluation}

\subsection{Training Dataset}
\label{sec:dataset}

Our training set consists of five difficult objects from the Amazon Robotics
Challenge 2017 (see \cref{fig:arc_objects}).
We selected objects that require precise grasping because they are heavy or unwieldy, and included
deformable and articulated objects.
For each object, we capture three sequences of 20 views, as described in \cref{sec:data_acq}.
For each view, we sample 60 new rotations along the camera axis and thus obtain 3600 training samples per object.
We artificially occlude portions of the image to make the network robust against occlusion that might occur in real-world scenarios.
The maximum occlusion percentage is limited to 50\,\%.
We split the dataset randomly into training and validation sets with a ratio of 80:20.

\subsection{Single-block Output vs. Multi-block Output}
\label{sec:eval:multiblock}

\begin{table}
\caption{Multi- vs. Single-block Output.}
\centering
\begin{threeparttable}
  \begin{tabular}{clcrr}
  \toprule
          &                                &                 & No occlusion & Occlusion\\
  \midrule
  \parbox[t]{2mm}{\multirow{4}{*}[-1mm]{\rotatebox[origin=c, ]{90}{Training}}} &
              \multirow{2}{*}{Translation [pix\tnote{1} ]} 
      & Single & 9.57& \textbf{11.21} \\
    & & Multi & \textbf{9.28} & 12.06 \\
    \cmidrule (lr){2-5}
    & \multirow{2}{*}{Orientation [$^\circ$]} 
    & Single & 5.92 & \textbf{6.44} \\
    & & Multi & \textbf{5.78}& 6.56 \\
    \midrule
  \parbox[t]{2mm}{\multirow{4}{*}[-1mm]{\rotatebox[origin=c]{90}{Validation}}} &
  \multirow{2}{*}{Translation [pix\tnote{1} ]} 
    & Single & 10.52 & \textbf{12.14} \\
    & & Multi & \textbf{9.68} &12.91 \\
    \cmidrule(lr){2-5}
    & \multirow{2}{*}{Orientation [$^\circ$]} 
    & Single & 7.9 & \textbf{9.76} \\
    & & Multi & \textbf{7.4}&9.64\\
    \bottomrule
  \end{tabular}
  \begin{tablenotes}
   \item Shown are translation and rotation errors on the validation set.
   \item [1] Relative to the 320$\times$320 input crop centered on the object.
  \end{tablenotes}
\end{threeparttable}
\label{tab:arch_comp}
\end{table}

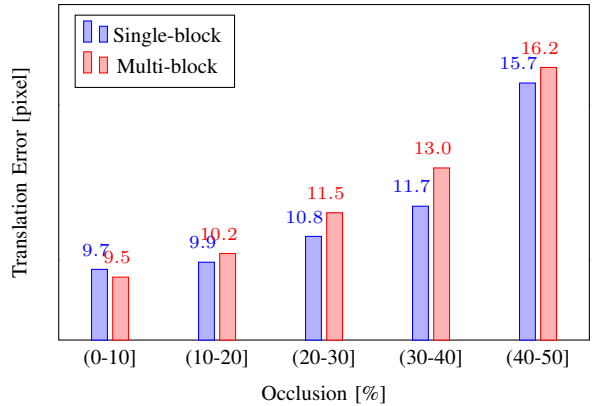
\begin{figure}
 \centering
 \begin{tikzpicture}
  \begin{axis}[
    ybar,
    bar width=6pt,
    font=\footnotesize,
yticklabels={,,},
    tickwidth         = 0pt,
    enlarge x limits  = 0.12,
    enlarge y limits  = 0.3,
    xtick={{(0-10]}, {(10-20]},{(20-30]}, {(30-40]}, {(40-50]} },
    xtick style={draw=none},
    symbolic x coords = {{(0-10]}, {(10-20]},{(20-30]}, {(30-40]}, {(40-50]} },
    nodes near coords,
legend pos=north west,
   ylabel={Translation Error [pixel]},
   xlabel={Occlusion [\%]},
   width=\linewidth,
   height=.7\linewidth,
   nodes near coords style={anchor=south east,inner sep=0,shift={(4pt, 5pt)},font=\scriptsize,/pgf/number format/.cd,fixed,fixed zerofill,precision=1}
  ]
  \addplot coordinates { 
({(0-10]}, 9.71)
({(10-20]},9.94)
({(20-30]},10.77)
({(30-40]},11.74)
({(40-50]},15.7) };

  \addplot coordinates { 

({(0-10]}, 9.46)
({(10-20]}, 10.22)
({(20-30]}, 11.53)
({(30-40]}, 12.97)
({(40-50]}, 16.2) };
  \legend{Single-block, Multi-block}
  \end{axis}
\end{tikzpicture}
 \caption{Effect of occlusion on the translation error. The images in the validation set 
 are grouped into different bins based on the percentage of pixels occluded. }
 \label{fig:xy_stats}
\end{figure}

\newlength{\sceneheight}
\setlength{\sceneheight}{2.7cm}

\begin{figure*}
 \centering
\includegraphics[height=\sceneheight]{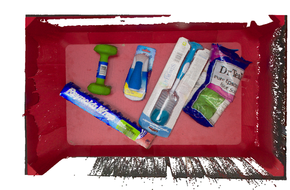}
\includegraphics[height=\sceneheight]{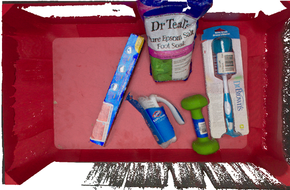}
\includegraphics[height=\sceneheight]{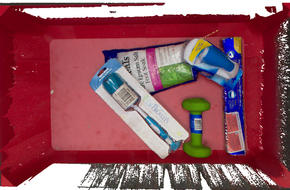}
\includegraphics[height=\sceneheight]{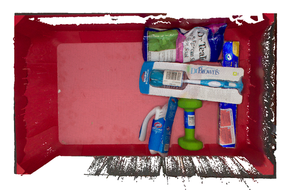}
 \vspace{1ex}
\includegraphics[height=\sceneheight]{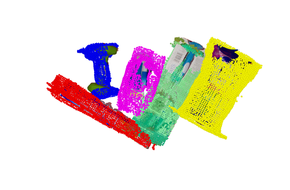}
\includegraphics[height=\sceneheight]{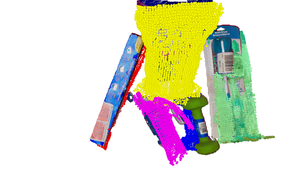}
\includegraphics[height=\sceneheight]{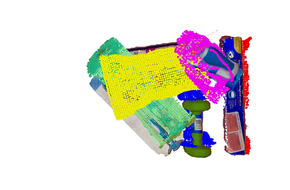}
\includegraphics[height=\sceneheight]{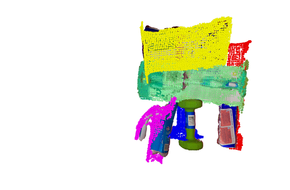}
 \vspace{1ex}
\includegraphics[height=\sceneheight]{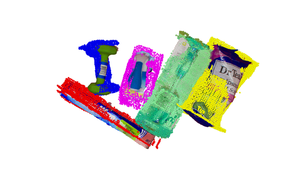}
\includegraphics[height=\sceneheight]{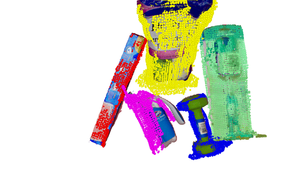}
\includegraphics[height=\sceneheight]{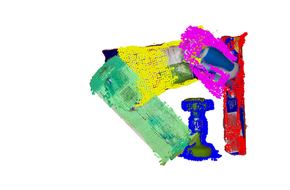}
\includegraphics[height=\sceneheight]{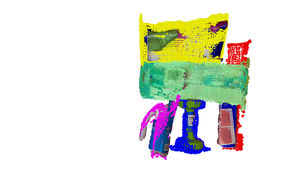}
 \caption{Experiments on complex scenes.
 Top row: Real cluttered tote scenes.
 Middle row: Object points extracted using semantic segmentation (shown in real colors)
 and predicted object poses visualized using object model clouds (in uniform colors).
 Bottom row: Same visualization after performing ICP refinement.}
 \label{fig:scenes}
\end{figure*}

We evaluated the performance of the single-block output and multi-block output variants of the pose estimation network on the synthetically generated dataset. The results of the comparison are provided in \cref{tab:arch_comp}. To understand the strengths and weaknesses of the architectures, we grouped the validation images into bins based on the percentage of pixels being occluded and analyzed the average error made by the variants in different bins.
\Cref{fig:xy_stats} shows the performance of both variants of the network under varying degrees of occlusion.
We observed that the multi-block output variant performs a little better in the absence of any occlusion, but the single-block output variant performs slightly better in the presence of occlusion.
The former can be explained by the fact that the training objective for the multi-block variant does not penalize for wrong object recognition; we simply discard the poses estimated in the blocks that do not correspond to the objects under consideration, i.e. we do not force the multi-block variant to perform object recognition as a part of pose estimation.
On the other hand, the single-block variant may be in less danger of overfitting,
resulting in better performance on occluded objects,
since it is forced to predict poses for different objects, much like a regularizer
introducing noise on the ground truth.

\begin{table}
\caption{Single-block Validation error on ARC objects.}
\centering
\begin{threeparttable}
  \begin{tabular}{lrrrr}
  \toprule
   Object & \multicolumn{2}{c}{Translation [pixel\tnote{1} ]} & \multicolumn{2}{c}{Orientation [$^\circ$]}    \\
   \cmidrule (lr) {2-3} \cmidrule (lr) {4-5}  
   & \hspace{2em}train & val                     & \hspace{1em}train & val      \\
  \midrule
  Browns brush     & 10.3  & 11.4  & 7.7 & 10.3   \\
  Epsom salts      & 11.2  & 12.5  & 7.4 & 10.5  \\
  Hand weight      &  9.6  & 10.4  & 2.1 & 2.6  \\
  Reynolds wrap    & 11.6  & 11.8  & 6.3 & 9.8  \\
  Utility brush    & 12.5  & 13.6  & 6.9 & 10.9  \\
  \bottomrule
  \end{tabular}
  \begin{tablenotes}
   \item [1] Relative to the 320$\times$320 crop centered on the object.
  \end{tablenotes}
\end{threeparttable}
\label{tab:arc_pose_val}
\end{table}

In all remaining experiments, we use the single-output variant. \Cref{tab:arc_pose_val} shows detailed quantified results on the validation set.

\subsection{Architecture Ablation Study.}

\begin{table}
\caption{Ablation study results on the synthetic dataset.}
\centering
\begin{threeparttable}
  \begin{tabular}{lrrrr}
  \toprule
  Model variant & \multicolumn{2}{c}{Translation [pixel]} & \multicolumn{2}{c}{Orientation [$^\circ$]}    \\
   \cmidrule (lr) {2-3} \cmidrule (lr) {4-5}  
   & \hspace{2em}train & val                     & \hspace{2em}train & val      \\
  \midrule
  FC per class           & 38.9  & 44.7  & 37.2  & 47.2  \\
  FC multi-class         & 46.4  & 54.4  & 42.7  & 51.9  \\
  1 Conv with stride 4   & 36.8  & 37.4  & 36.3  & 44.5  \\
  1 Conv with stride 8   & 36.4  & 37.1  & 25.8  & 34.2  \\
  2 Conv with stride 2   & 32.3  & 33.9  & 11.3 & 17.0 \\
  3 Conv with stride 2   & 10.2  & 13.5  & 4.64  & 10.8  \\
  \bottomrule
  \end{tabular}
\end{threeparttable}
\label{tab:ablation}
\end{table}

Our design of the pose estimation network architecture is motived by the need to run the pose estimation after a semantic segmentation network using the same backbone network.
Thus, we wanted the pose estimation architecture to be as light as possible.

We created individual linear models for each object class separately which we considered as the baseline to evaluate the performance of different
architectural designs. 
The linear model consisted of just one fully connected layer on top the 80$\times$80$\times$256 feature maps to regress the 5D pose.
The results presented in the first row of \cref{tab:ablation} indicate that this model cannot learn the task.

In a multi-class setting, the fully connected model performs even worse (see \cref{tab:ablation}).
Adding more fully connected hidden layers is not feasible due to GPU memory constraints
caused by the large size of the input feature maps, which limits the number of hidden neurons.
Thus, the architectures that do not reduce the input feature dimension failed to attain adequate performance.
This leaves us with two direct options to reduce the size of input dimension: using pooling layers, or using convolutional layers with \mbox{stride $>1$}.
Here, we use convolutional layers with 256 features and \mbox{stride $>1$}.
Results from an evaluation of different architectures are presented in \cref{tab:ablation}.
We identified the best architecture to be three dimension-reducing convolutional layers of stride 2.
From our perspective, the results indicate that a certain capacity of the model is needed
to learn the pose estimation tasks and that convolutional architectures are well-suited for this purpose.

\subsection{Complex Scenes}

\begin{table}
\caption{Results on complex scenes.}\label{tab:arc_pose_test}
\centering
  \begin{tabular}{lrrrr}
  \toprule
  \multirow{2}{*}{Object} & \multicolumn{2}{c}{Translation [pixel]} & \multicolumn{2}{c}{Orientation [$^\circ$]}    \\
  \cmidrule (lr) {2-3} \cmidrule(lr){4-5}
  & Predicted & Refined & Predicted & Refined  \\
  \midrule
  Browns brush     & 12.75  & 12.14 & 14.63 & 14.26 \\
  Epsom salts      & 14.43  & 16.12 & 16.45 & 15.27 \\
  Hand weight      & 16.18  & 15.42 & 10.34 & 9.73 \\
  Reynolds wrap    & 12.78  & 12.48 & 18.37 & 16.04 \\
  Utility brush    & 16.74  & 15.87 & 15.75 & 14.88 \\
  \bottomrule
  \end{tabular}
\end{table}

We collected 14 real scenes as they might have occurred during the Amazon
Robotics Challenge 2017\footnote{The complex scenes dataset is publicly available at \mbox{\url{http://www.ais.uni-bonn.de/data/pose_estimation/}}}. 
They contain the five objects in varying levels of
occlusion and were manually annotated with ground truth poses.
Examples of the scenes are shown in the top row of \cref{fig:scenes}.
The five objects possess different physical properties ranging from shiny
surfaces in the case of the reynolds wrap to a symmetrical dumbbell and a
deformable salt bag.
The predicted 6D poses are overlaid with points of uniform color shown in the
middle row of \cref{fig:scenes}.
In general, we can observe that the 6D pose predicted by our method is acceptable.
The ability of the model to handle occlusion and undersegmentation is demonstrated in the third column where the objects are lying on top of the other objects and the last column where a portion of the reynolds wrap object is undersegmentated.
Using Iterative Closest Point (ICP) refinement, the result can be further improved.
The final pose after performing Generalized-ICP (GICP)~\citep{segal2009generalized} is shown in the bottom row of \cref{fig:scenes}. 
In computing the orientation error for symmetrical objects, for example, along $yaw$ axis for the hand weight object, we ignore the error along the symmetrical axis by the following steps:
\begin{enumerate}

 \item computing the error between the ground truth and the predicted quaternion $q_{error}$,
 \item extracting the angular component $\Psi$ of the $q_{error}$ along the axis of the symmetry,
 \item creating a new quaternion representing the rotation by $\Psi$ along the symmetrical axis $q'$, and
 \item premultiplying $q_{error}$ with the computed $q'$ and thus rotating out the error along the symmetrical axis.
\end{enumerate}

\subsection{Generalization}

\begin{table}
\caption{Generalization error.}\label{tab:general}
\centering%
  \begin{tabular}{lrr}
   \toprule
                & Translation [pixel] & Rotation [$^\circ$] \\
   \midrule
   No occlusion & 36.34 & 33.60 \\
   Occlusion    & 39.52 & 38.21 \\
   \bottomrule
  \end{tabular}
\end{table}

\begin{figure}
 \centering
 \adjustbox{valign=c}{\begin{minipage}{.5\linewidth}
 \includegraphics[width=0.45\linewidth,clip,trim={112 18 112 87}]{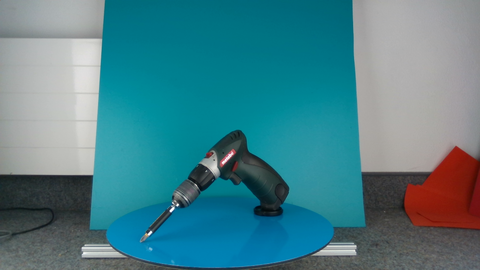}
 \includegraphics[width=0.45\linewidth,clip,trim={112 18 112 87}]{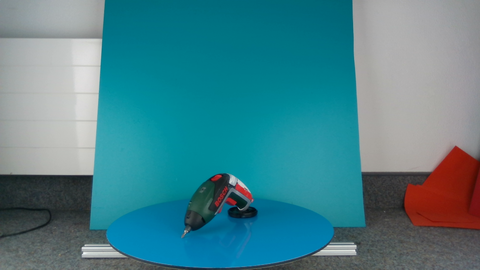}

 \vspace{4pt}

 \includegraphics[width=0.45\linewidth,clip,trim={112 18 112 87}]{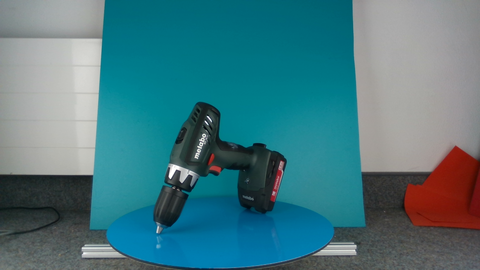}
 \includegraphics[width=0.45\linewidth,clip,trim={112 18 112 87}]{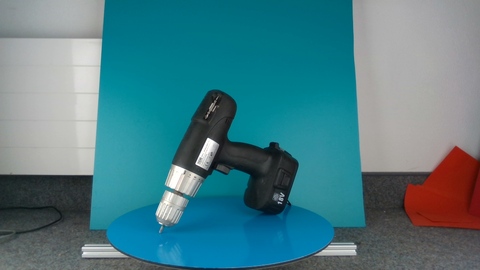}
 \end{minipage}}\hfill
 \adjustbox{valign=c}{\includegraphics[width=0.4\linewidth,clip,trim={112 18 112 87}]{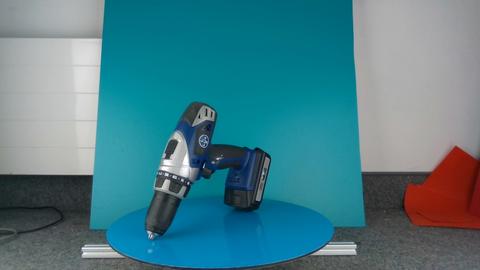}}
  \caption{Objects for the generalization experiment.
  Left: Training objects.
  Right: Test object.}
\label{fig:drills}
\end{figure}

\begin{figure}
 \centering\setlength{\fboxsep}{0pt}\newlength{\genwidth}\setlength{\genwidth}{0.18\linewidth}
  \fbox{\includegraphics[width=\genwidth,clip,trim={325 110 325 110}]{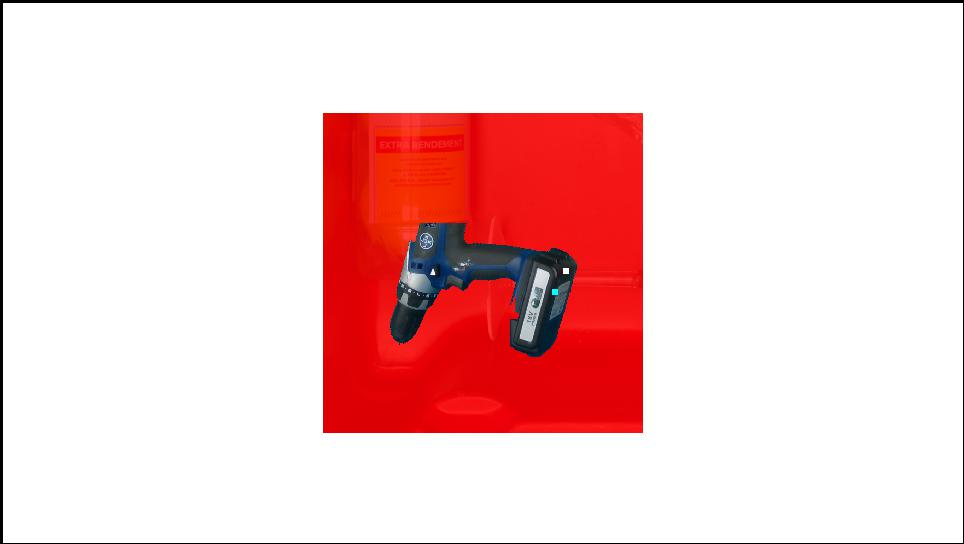}}\hfill
  \fbox{\includegraphics[width=\genwidth,clip,trim={325 110 325 110}]{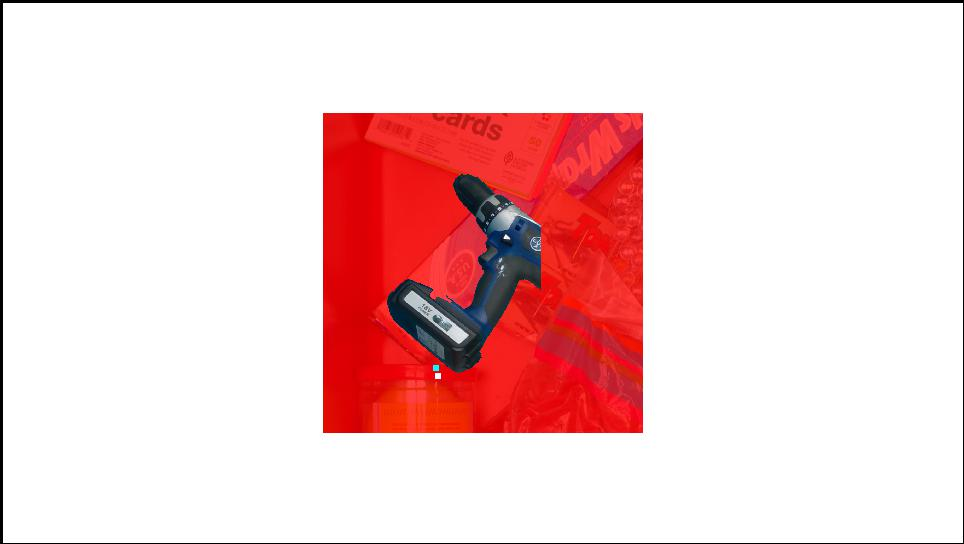}}\hfill
  \fbox{\includegraphics[width=\genwidth,clip,trim={325 110 325 110}]{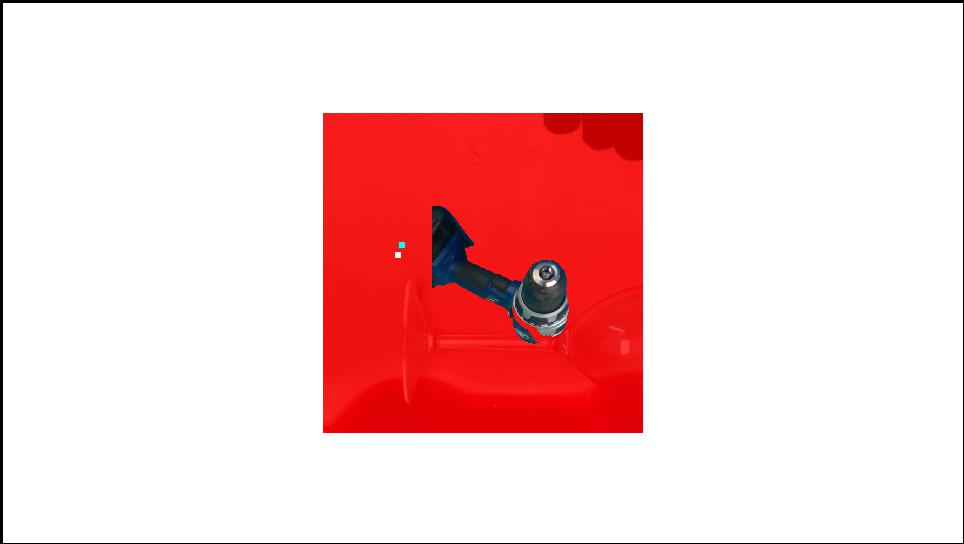}}\hspace{0.3cm}
  \fbox{\includegraphics[width=\genwidth,clip,trim={325 110 325 110}]{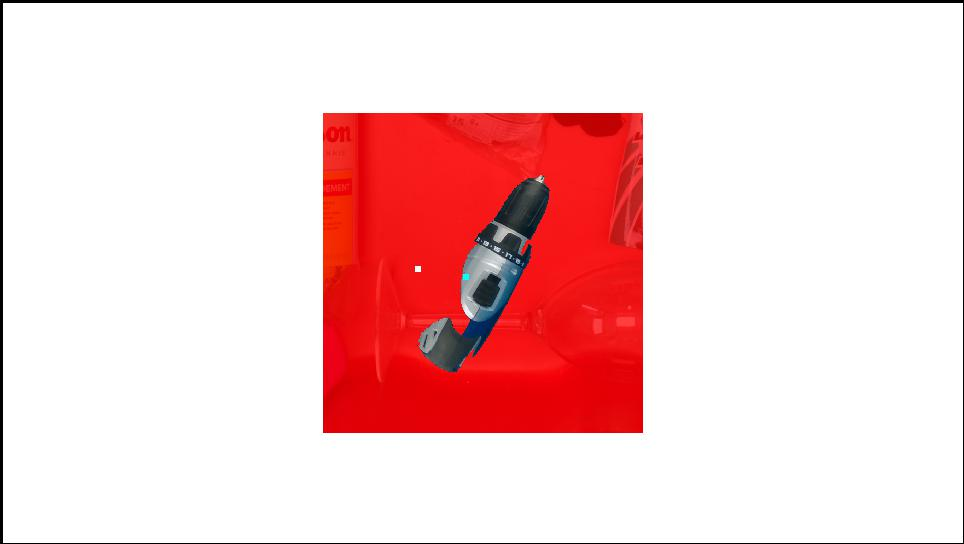}}\hfill
  \fbox{\includegraphics[width=\genwidth,clip,trim={325 110 325 110}]{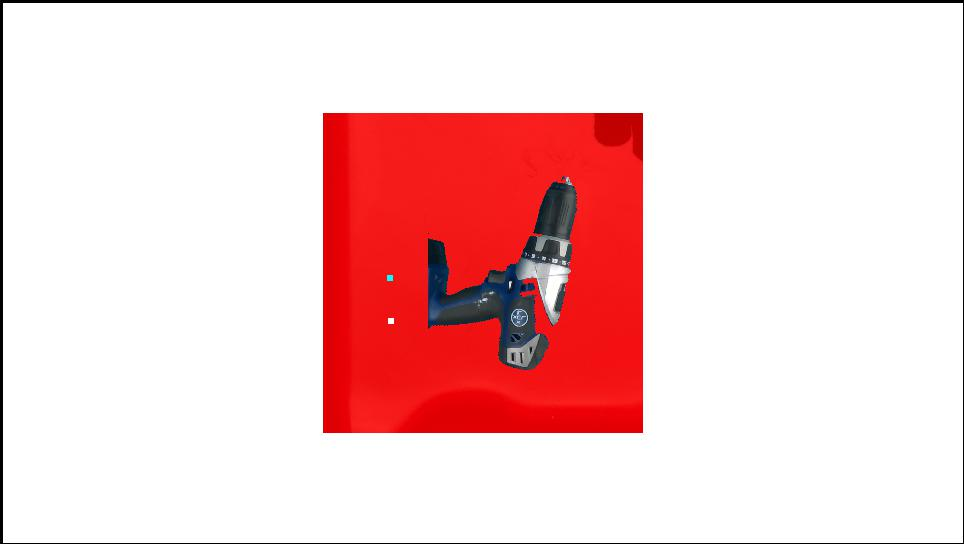}}

  \vspace{.5ex}

  \fbox{\includegraphics[width=\genwidth,clip,trim={355  80 295 140}]{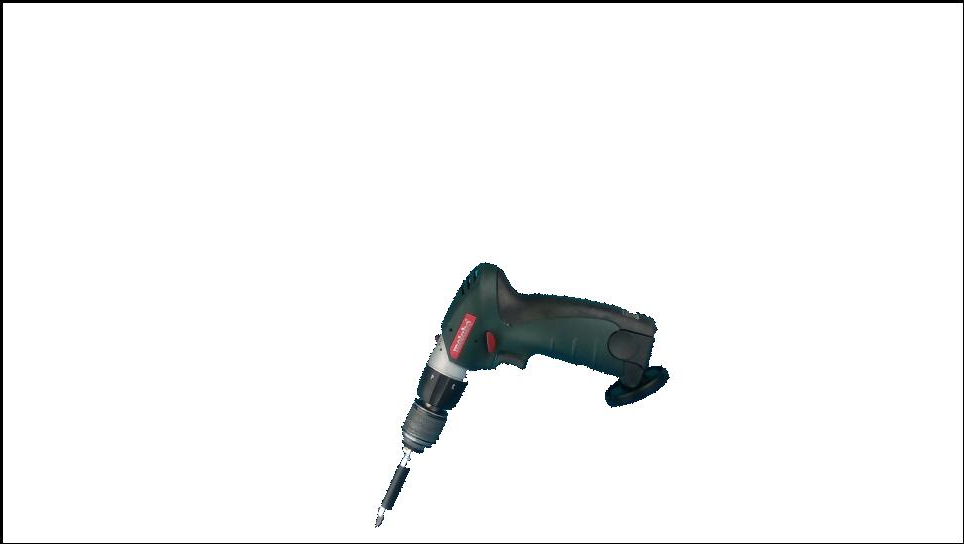}}\hfill
  \fbox{\includegraphics[width=\genwidth,clip,trim={265 180 385  40}]{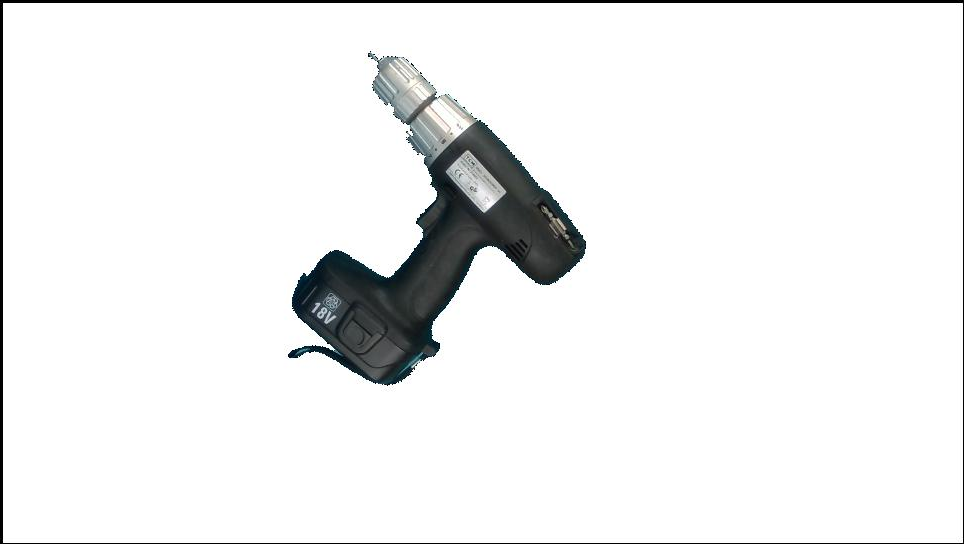}}\hfill
  \fbox{\includegraphics[width=\genwidth,clip,trim={400 180 250  40}]{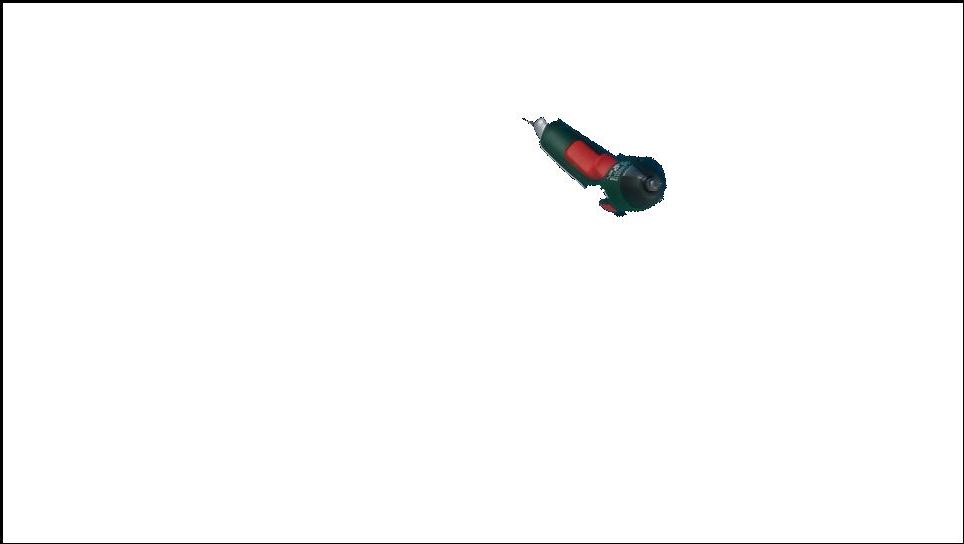}}\hspace{0.3cm}
  \fbox{\includegraphics[width=\genwidth,clip,trim={415 150 235 70}] {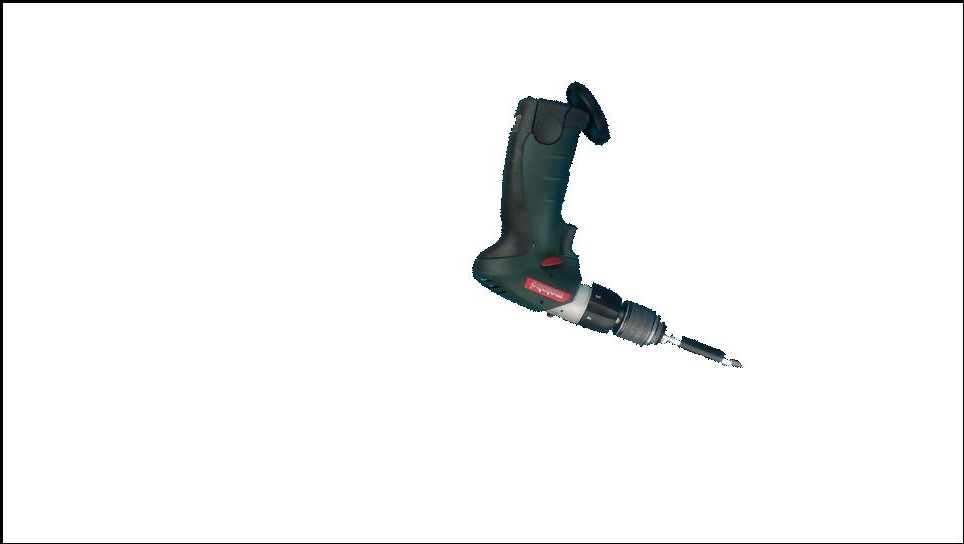}}\hfill
  \fbox{\includegraphics[width=\genwidth,clip,trim={240 210 410   10}]{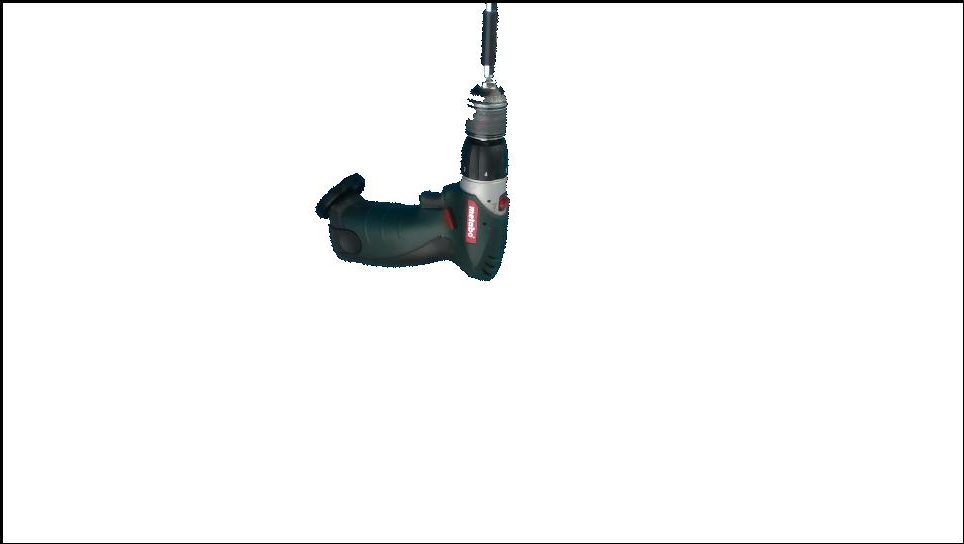}}
  \caption{Generalization with occlusion;
  Top row: Input image to the network. White and green dots show ground truth and predicted object centers, respectively.
  Bottom row: Image in the dataset closest to the orientation predicted.
  The first three images show working cases, while the last two show typical failure cases in orientation and translation, respectively.}
\label{fig:general_occ_working}
\end{figure}

The objects that we encounter in the real world are often instances of some object category.
Humans do not need to learn to recognize/operate each single instance of the same category,
instead transferring the knowledge acquired for some instances to others of the same category.
With enough training data, CNNs have been shown to generalize to new instances of same category.
In this experiment, we evaluate the generalization properties of the pose estimation networks.
We trained the pose estimation networks on a set of four drills, evaluating
the performance on a different drill (see \cref{fig:drills}).
The accuracy of the pose estimation network in predicting the pose of the unseen drill with and without occlusion is shown in \cref{tab:general}.
We investigated the pose estimation for the unseen cases further by retrieving the image in the training set with the closest orientation to the predicted one. This allows direct visual comparison of the predicted poses (see \cref{fig:general_occ_working} for typical working and failure cases).
In the last image image of \cref{fig:general_occ_working}, while the predicted orientation is comparatively good, the translation prediction is off by a wider margin due the bottom portion of the drill being occluded.

\subsection{Symmetries}

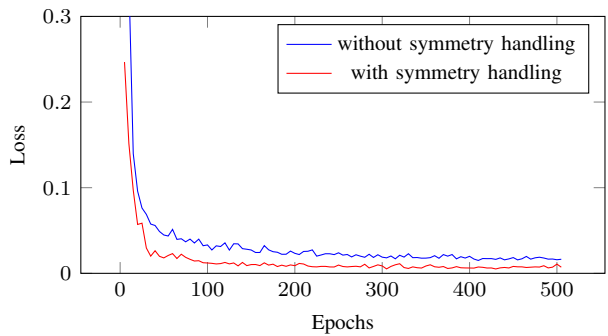
\begin{figure}
 \centering
    \begin{tikzpicture}[font=\footnotesize]
      \begin{axis} [width=.99\linewidth, height=5cm, ymin=0.0, ymax=0.3, xlabel=Epochs, ylabel=Loss, legend pos = north east, no markers]
	  \addplot table [col sep=comma,x expr=\thisrow{Idx}*5,y={hand weight}] {images/symmetry/inv_non/objectvalidationqnormloss};
	  \addplot table [col sep=comma,x expr=\thisrow{Idx}*5,y={hand weight inv}]   {images/symmetry/inv_inv/objectvalidationqnormloss};

	  \addlegendentry{{without symmetry handling}}
	  \addlegendentry{{with symmetry handling}}
      \end{axis}
    \end{tikzpicture}
\caption{Effect of symmetry handling on the learning curve.}
\label{fig:inv_non}
\end{figure}

We quantified the benefits of addressing the symmetry in the pose of the dumbbell by analyzing the loss during the training process as shown in \cref{fig:inv_non}.
In the case of not handling symmetries, the model needs around 300 epochs to achieve a reasonable performance compared to the final convergence whereas in the case of handling the symmetries, it needs only around 100 epochs. Thus, the speed of convergence is significantly faster when the symmetry in the poses is handled. The network also converged to a better accuracy.
This demonstrates the advantages of dealing with the symmetry in the poses.

\subsection{Limitations}

One major weakness of our automatic data acquisition pipeline are deformable or articulated objects.
Our data capture setup is optimized for fast data acquisition and needs only two or three sequences of turntable captures per object. We apply randomly sampled rotations around the camera axis to synthesize object appearances in new poses. But in the case of deformable or articulated objects,
appearance will be affected by gravity or collisions, which is not captured by our data
augmentation step.
Also, our method relies heavily on the performance of the semantic segmentation module.
While undersegmentation is to some degree is modeled as occlusion during training, oversegmentation can affect the performance of pose estimation.

\section{Conclusion}
We presented a pipeline for 6D object pose estimation in clutter, designed for scenarios where the new objects are to be learned quickly. Our pipeline consists of 
\begin{enumerate}
 \item a fast data acquisition and synthetic training data generation module that needs minimal human intervention,
 \item a semantic segmentation module to segment the cluttered scene, and 
 \item an object pose estimation module that regresses the 5D pose of the objects from the segmented RGB crop of the scene.
\end{enumerate}
We presented and compared two CNN architectures for pose estimation from RGB images and analyzed their strengths and weaknesses in their ability to deal with occlusion. Our method performs semantic segmentation of the scene to extract the crops of objects and estimates the 5D pose of the object: orientation and translation in 2D image plane. The 2D translation prediction in the image plane is projected into 3D using depth information. We also proposed a method to deal with symmetries---similarities on the object appearance under different poses---by assigning the same ground-truth pose for all the poses that exhibit minimal variation in their appearance. Finally, we demonstrated the usability of our method on complex bin-picking scenarios and discussed the performance and limitations of our method.

% \printbibliography
\bibliographystyle{IEEEtranN}
\bibliography{references}

\end{document}